\documentclass[sigconf]{acmart}

\usepackage{amsmath}
\usepackage{amsthm}
\usepackage{balance}

\usepackage{graphicx}
\usepackage{multirow}
\usepackage{subfigure}
\usepackage{amsmath}
\usepackage[linesnumbered,ruled]{algorithm2e}

\newcommand{\ourmethod}{Robust-PU\xspace}

\AtBeginDocument{%
  \providecommand\BibTeX{{%
    \normalfont B\kern-0.5em{\scshape i\kern-0.25em b}\kern-0.8em\TeX}}}

\copyrightyear{2023} 
\acmYear{2023} 
\setcopyright{acmlicensed}\acmConference[KDD '23]{Proceedings of the 29th ACM SIGKDD Conference on Knowledge Discovery and Data Mining}{August 6--10, 2023}{Long Beach, CA, USA}
\acmBooktitle{Proceedings of the 29th ACM SIGKDD Conference on Knowledge Discovery and Data Mining (KDD '23), August 6--10, 2023, Long Beach, CA, USA}
\acmPrice{15.00}
\acmDOI{10.1145/3580305.3599491}
\acmISBN{979-8-4007-0103-0/23/08}

\begin{document}

\title{Robust Positive-Unlabeled Learning via Noise Negative Sample Self-correction}

\author{Zhangchi Zhu}
\authornote{Work done at Microsoft Research Asia.}
\email{zczhu@stu.ecnu.edu.cn}
\orcid{0009-0007-7643-3717}
\affiliation{%
  \institution{East China Normal University}
  \city{Shanghai}
  \country{China}
}

\author{Lu Wang}
\orcid{0000-0002-7305-1496}
\email{wlu@microsoft.com}
\affiliation{
  \institution{Microsoft Research}
  \city{Beijing}
  \country{China}}

\author{Pu Zhao}
\orcid{0000-0002-4518-323X}
\email{pu.zhao@microsoft.com}
\affiliation{
  \institution{Microsoft Research}
  \city{Beijing}
  \country{China}}

\author{Chao Du}
\orcid{0009-0008-2893-5461}
\email{chaodu@microsoft.com}
\affiliation{
  \institution{Microsoft Research}
  \city{Beijing}
  \country{China}}

\author{Wei Zhang}
\orcid{0000-0001-6763-8146}
\email{zhangwei.thu2011@gmail.com}
\authornote{Corresponding authors.}
\affiliation{
  \institution{East China Normal University}
  \city{Shanghai}
  \country{China}}

\author{Hang Dong}
\orcid{0000-0001-6439-8183}
\email{hangdong@microsoft.com}
\affiliation{
  \institution{Microsoft Research}
  \city{Beijing}
  \country{China}}

\author{Bo Qiao}
\orcid{0000-0002-8997-8317}
\email{boqiao@microsoft.com}
\affiliation{
  \institution{Microsoft Research}
  \city{Beijing}
  \country{China}}

\author{Qingwei Lin}
\orcid{0000-0003-2559-2383}
\authornotemark[2]
\email{qlin@microsoft.com}
\affiliation{
  \institution{Microsoft Research}
  \city{Beijing}
  \country{China}}

\author{Saravan Rajmohan}
\orcid{0000-0002-2019-213X}
\email{saravan@microsoft.com}
\affiliation{
  \institution{Microsoft 365}
  \city{Seattle}
  \country{USA}}

\author{Dongmei Zhang}
\orcid{0000-0002-9230-2799}
\email{dongmeiz@microsoft.com}
\affiliation{
  \institution{Microsoft Research}
  \city{Beijing}
  \country{China}}

\renewcommand{\shortauthors}{Zhangchi Zhu, et al.}

\begin{abstract}
Learning from positive and unlabeled data is known as positive-unlabeled (PU) learning in literature and has attracted much attention in recent years. One common approach in PU learning is to sample a set of pseudo-negatives from the unlabeled data using ad-hoc thresholds so that conventional supervised methods can be applied with both positive and negative samples. Owing to the label uncertainty among the unlabeled data, errors of misclassifying unlabeled positive samples as negative samples inevitably appear and may even accumulate during the training processes. Those errors often lead to performance degradation and model instability. To mitigate the impact of label uncertainty and improve the robustness of learning with positive and unlabeled data, we propose a new robust PU learning method with a training strategy motivated by the nature of human learning: easy cases should be learned first. Similar intuition has been utilized in curriculum learning to only use easier cases in the early stage of training before introducing more complex cases. Specifically, we utilize a novel ``hardness'' measure to distinguish unlabeled samples with a high chance of being negative from unlabeled samples with large label noise. An iterative training strategy is then implemented to fine-tune the selection of negative samples during the training process in an iterative manner to include more ``easy'' samples in the early stage of training. Extensive experimental validations over a wide range of learning tasks show that this approach can effectively improve the accuracy and stability of learning with positive and unlabeled data. Our code is available at \url{https://github.com/woriazzc/Robust-PU}.
\end{abstract}




\maketitle

\section{Introduction}

Conventional supervised binary classification problems often assume that all training samples are clearly labeled as positive (P) data and negative (N) data. However, correctly labeling all positive samples in the training data can be costly or impractical in many real applications, including images classification ~\cite{geurts2011learning} and priority ranking~\cite{mordelet2011prodige,mordelet2014bagging} of genes or gene combinations associated with diseases. It is then often the case that only a relatively small amount of positive data is reliably labeled and available for training along with a large set of unlabeled (U) data. Learning with such data is known as Positive-Unlabeled (PU) learning and has attracted much attention in recent years. 


While PU learning problems can be solved by  conventional supervised learning approaches by treating all the unlabeled samples as negative samples, the contamination of positive samples among unlabeled samples would introduce considerable bias to the learning process \cite{bekker2020learning}. To address this issue, existing PU learning methods usually adopt one of the following two strategies: (1) Sample selection strategy, in which negative samples are identified from the unlabeled data for training. Usually, the trained classification models are used to generate and update the negative labels based on ad-hoc thresholds at each training step in a recursive fashion. However, during the early training stages, the poorly-trained classification models can easily misclassify unlabeled samples especially when the prior of positive samples (proportion of positive samples among unlabeled data) is relatively high. Such misclassification errors could accumulate and cause persistent bias and instability~\cite{Xu2019RevisitingSS} in the learning process; (2) De-biasing strategy, in which new classification risks are developed to support unbiased learning with positive and unlabeled data. This strategy, while circumvents the issue of labeling the unlabeled data, requires the knowledge of the prior of positive samples for constructing the unbiased risks, which can be difficult to estimate accurately in practice. 


Despite all those developments, recent studies~\cite{xu2019revisiting} show that many existing PU learning methods are still susceptible to noise in the unlabeled data and tend to overfit the noisy negatives. In this paper, we propose a robust PU learning method with a new training strategy to alleviate this issue. This strategy is motivated by the nature of human learning: it is often better to learn the easier knowledge first before exposing it to the harder one. This ``easy-to-hard'' strategy has been used as one of the most popular approaches of curriculum learning \cite{wang2021survey}, in which only the easier concepts (i.e. recognizing objects in simple scenes with clearly visible objects) are used for training the model during the early stage of learning, while more complex cases such as cluttered images with occlusions are only introduced later. It has been verified that such an ``easy-to-hard'' strategy can help learn more robust models in scenarios with noisy data.

In particular, our new training strategy employs an iterative approach for fine-tuning the selection of negative samples throughout the training process. In the early stage of training, only the ``easy'' samples, unlabeled samples with a very high chance of being negative, are selected for training the classification model. More samples, including ``hard'' ones with higher labeling noise, are introduced in a gradual fashion as the training processes. To support this strategy, we develop a novel hardness measurement based on the classification loss to quantify the degree of noise in unlabeled samples. Furthermore, we implement a training scheduler to oversee the sample selection process. This scheduler utilizes a dynamic weight-based threshold for selecting negative samples from unlabeled data at each iteration. As the number of iterations increases, this threshold would be smoothly relaxed to ensure the inclusion of more samples in the later stages of training. Our experiments show that this strategy can reliably improve the robustness and generalization of the trained models across a wide range of PU learning tasks.   





In summary, we make the following contributions:
\begin{itemize}
\item We demonstrate that a well-designed training strategy based on the ``from easy to hard'' principle can lead to considerable improvement in PU learning methods and suggest new research opportunities in this direction. 

\item We propose a new difficulty measurement for measuring the ``hardness'' of PU data and explore a set of training schedulers for PU learning. 

\item We propose a novel training strategy called \ourmethod based on the ``hardness'' measure and the training scheduler. By dynamically increasing the expected number of selected clean negatives in the training process, this strategy effectively reduces the influence of noisy negatives and significantly improves the generalization and robustness ability of the trained model in PU learning tasks. 
\item We conduct extensive experimental validations on the effectiveness of the proposed method over a wide range of PU learning tasks.
\end{itemize}

\begin{figure*}[htbp]
\centering
  \includegraphics[width=0.84\linewidth]{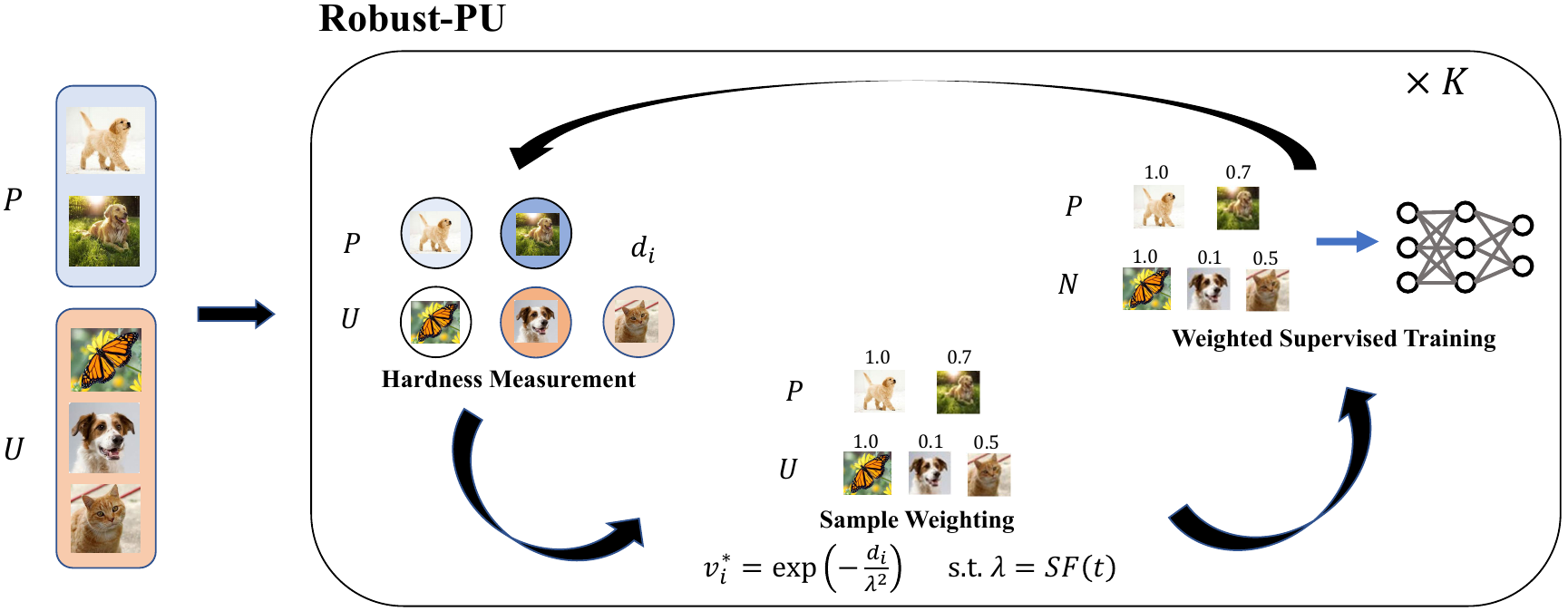}
  \caption{Overview of the proposed \ourmethod. We propose a novel iterative training strategy to train a classification model with only positive and unlabeled data. We divide each iteration into three stages: 1) Hardness Measurement, 2) Sample Weighting, and 3) Weighted Supervised Training. By updating the weights of both positive and unlabeled samples dynamically regarding their ``hardness'', we can greatly alleviate the problem of overfitting the noisy negatives.}
  \label{fig:pipline}
\end{figure*}

\section{Related Works}

\subsection{Positive-Unlabeled Learning}Existing PU learning methods mainly fall into two categories. The first branch relies on negative sampling from unlabeled data and is also known as biased PU learning. These methods~\cite{liu2002partially,yu2004pebl} firstly identify reliable negative samples from unlabeled data and then apply a regular supervised learning strategy with positive and negative data. PUbN~\cite{hsieh2019classification} uses the pre-trained model to recognize a small portion of negative samples for calculating risks associated with positive, negative, and unlabeled samples. PULNS~\cite{luo2021pulns} combines PU learning with reinforcement learning. It uses an agent to select negative samples and treats the binary classifier's performance on the validation dataset as the reward. Except for selecting negative samples from the real dataset, GenPU~\cite{hou2017generative} also adopts the GAN framework to generate extra data as the input of the classifier. Still, at the early stage of training, the poorly-trained classification model can easily misclassify unlabeled samples. Such misclassification errors could accumulate and cause persistent bias and instability in the learning process. 

The second category of PU methods adopts the framework of cost-sensitive learning, in which new classification risks~\cite{du2014analysis} are developed to support unbiased classification with positive and unlabeled data directly. Due to the strong flexibility of deep neural networks, empirical risks on training data may go negative and lead to serious overfitting. This issue is addressed by nnPU~\cite{kiryo2017positive} that bounds the estimation of the risk on negative data. Self-PU~\cite{chen2020self} proposes to use the unlabeled samples via a self-learning method including model calibration and distillation. It is worth noting that although Self-PU tries to solve the problem via self-paced learning, it only takes it as a warm-up and fails to utilize the full power of the easy-to-hard principle, resulting in an insufficient use of true negatives. Dist-PU\cite{zhao2022dist} aligns the expectations of the predicted and true label distributions to improve the label consistency and adopts entropy minimization and Mixup regularization to avoid the trivial solution. All these methods assume that the labeled positive data is identically distributed as the unlabeled positive, which is unrealistic in many instances of PU learning. To this end, PUSB~\cite{kato2018learning} takes a mild assumption that preserves the order induced by the class posterior. In general, while all those methods provide means for learning unbiased risk estimators from unlabeled data, they all assume that the prior distribution of positive samples is known or can be accurately estimated. There are also some studies aiming at estimating the class prior for PU learning. PE~\cite{christoffel2016class} uses penalized divergences for model fitting to cancel out the error caused by the absence of negative samples. CAPU~\cite{chang2021positive} jointly optimizes the prior estimation and the binary classifier by pursuing the optimal solution of gradient thresholding. Still, estimating class prior remains a difficult problem which is often harder than the classification task itself.

\subsection{Curriculum Learning}
Curriculum learning (CL) is inspired by human learning. Early curriculum learning studies~\cite{khan2011humans,basu2013teaching,zhou2021curriculum,spitkovsky2009baby} seek an optimized sequence of training samples (i.e. a curriculum, which can be designed by human experts) to improve model performance. The main idea of curriculum learning is to ``train from easier data to harder data''. In practice, CL usually assigns greater weights to easy samples and gradually increases the weights of each sample in the dataset, eventually using the full dataset for training. As a general strategy, curriculum learning has been successfully applied to a variety of tasks, including computer vision (CV)~\cite{guo2018curriculumnet}, natural language processing (NLP)~\cite{platanios2019competence}, recommender system (RS)~\cite{WangZLLZLZ23}, etc.

Many studies adopt CL for denoising and validate its effectiveness in improving the robustness of models. Specifically, they use samples with high confidence to alleviate the interference of noisy data from the perspective of data distribution. Previous works~\cite{zhou2021robust} have validated that CL can help make the training more robust and more generalizable. The most popular example of using CL for denoising is combining CL with neural machine translation (NMT)~\cite{kumar2019reinforcement}, which always uses a highly heterogeneous and noisy dataset. CL is also used in weakly-supervised CV tasks to select clean samples from noisy data collected from the web~\cite{guo2018curriculumnet}.

Self-paced learning (SPL) is a branch of CL that takes the training loss given by the current model as the difficulty of samples. It introduces a regularizer and alternatively optimizes the loss and regularizer. Then the weight of samples is given by the minimizer function of the regularizer. The original SPL~\cite{kumar2010self} uses regularizers that lead to binary weights (i.e. 0 or 1). However, they ignore the difference between samples with the same weights. Therefore, an intuitive choice is to design new SP-regularizers to result in soft weights~\cite{jiang2014easy}. Except for the regularizers that have explicit forms, Fan et al.~\cite{fan2017self} introduced implicit regularizers, whose analytic forms are unknown.
They are deduced from some well-studied robust loss functions, and their minimizer functions can be derived from these loss functions.

Our work shares a similar intuition with CL methods. However, it is not trivial to apply CL to PU learning. This is because, in PU learning, unlabeled data usually occupies the vast majority and only a small portion of the positive samples are labeled.
As a result, the training process can be easily interfered with by the unlabeled positives in unlabeled samples. To ensure that PU learning can benefit from similar training strategies as those used in CL methods, a new pipeline with both hardness measurement and training scheduler suitable for both positive and unlabeled data is needed.


\section{Method}

In this section, we will introduce our training strategy in detail. Our new training strategy converts semi-supervised training on PU data into weighted supervised training with dynamic sample weights and contains three iterative stages. In Section~\ref{sec:overview}, we will explain the role of each stage. Then, in each of the three subsequent sections, we will detail a corresponding stage and the new components we proposed for that stage. Finally, we will provide a holistic view of our method in Section~\ref{sec:overall}.

\subsection{Problem Setup}
Here we first introduce the notations in PU learning.
Let $X\in \mathbb{R}^d, Y\in\{0,1\}$ be the input random variables and the output variable, respectively. In PU learning, the training set $D$ consists of two parts, the positive data $\mathcal{X}_p$ and the unlabeled data $\mathcal{X}_u$. $\mathcal{X}_p$ contains $n_p$ instances sampled from $P(X|Y=1)$. $\mathcal{X}_u$ contains $n_u$ instances sampled from $P(X)$. The goal of PU learning is to learn a model $f_\omega(x_i)$ with the optimized parameter $\omega$ to map instance $x_i$ into the predicted label $\hat{y}_i$.

\begin{algorithm}
    \caption{\ourmethod}\label{alg:pu}
    \SetKwInOut{Input}{Input}
    \SetKwInOut{Output}{Output}
    \Input{Training data $(\mathcal{X}_p, \mathcal{X}_u)$;\\
    Iteration number $T$; Epoch number $E$.}
    \Output{Model Parameter $\theta$.}
    Initialize the model by pre-training using nnPU.\\
    \For{$t=1$ to $T$}
    {
    \textbf{Stage 1: Hardness Measurement}\\
        Measure the hardness of each sample in $\mathcal{X}_p$ and $\mathcal{X}_u$ according to Eq. (\ref{eq:log}) or (\ref{eq:sig}).\\
    \textbf{Stage 2: Sample Weighting}\\
    Calculate the threshold $\lambda=F(t)$ according to one of the pacing functions in Eq. (\ref{sf:linear}) \textasciitilde (\ref{sf:exp}).\\
    Update the weight $v^*_i$ of each sample according to Eq. (\ref{eq:wei}).\\
    \textbf{Stage 3: Weighted Supervised Training}\\
    \For{epoch $i=1$ to $E$}
    {
    Set the ground truth label of unlabeled data as 0.
    Update the model parameter $\theta$ by minimizing the loss Eq. (\ref{eq:bce}).
    }
    }
\end{algorithm}

\subsection{Overview}\label{sec:overview}
As shown in Figure~\ref{fig:pipline}, we design an iterative training strategy for selecting easy positive samples from the positive dataset and reliable negative samples from the unlabeled dataset to enhance the model’s ability to distinguish the positive and negative samples.
Specifically, we initialize the model by pre-training using nnPU to speed up the training. Then we progressively increase the difficulty of the negative candidates associated with the positive samples in the training set. In this way, the model is encouraged to gradually distinguish the positive and negative samples.
We describe our algorithm \ourmethod in Alg.~\ref{alg:pu}.

We divide each iteration into three stages:

\begin{itemize}
\item In the hardness measurement stage, we use a ``hardness'' metric function to recognize reliable negative data from unlabeled data and determine easy samples from positive data.
\item In the sample weighting stage, we update the weight of each sample based on its ``hardness'' and a changing threshold. In order to realize learning from easy to hard and from clean to noisy, we propose several schedulers to change the threshold.
\item In the final stage, we regard the unlabeled data as negative and convert the original semi-supervised training into supervised training with weighted samples.
\end{itemize}


\subsection{Hardness Measurement}

The difficulty of PU learning is twofold: differentiation and noisiness. Differentiation indicates if the positive samples can be differentiated from the negative samples. Noisiness indicates if the selected pseudo-negatives are true negatives. Thus we measure the hardness for both positive and negative samples.

To evaluate the differentiation and noisiness, the classification loss has been widely used~\cite{xu2019revisiting}, which gives low (high) differentiable and clean samples a high (low) loss. Specifically, we calculate the classification loss using +1 as the ground truth for positive samples. And for unlabeled samples, in this stage, we treat all of them as negative samples and calculate the loss using -1 as the ground truth. Thus we formulate the key idea of hardness with two functions, as we called the hardness metric function $d$:

\begin{itemize}
    \item Logistic loss
\begin{align}
    d_i = \log(1+\exp(-\tilde{y_i}z_i/\tau))\label{eq:log}
\end{align}
where $z$ denotes the logit, $\tau$ denotes the temperature, and $\tilde{y}$ denotes the ``pseudo ground-truth'', which is +1 for positive samples and -1 for unlabeled samples.
\item Sigmoid loss 
\begin{align}
    d_i = \frac{exp(-\tilde{y_i}z_i/\tau)}{1+exp(-\tilde{y_i}z_i/\tau)}\label{eq:sig}
\end{align}
\end{itemize}


 



In the above two functions, we adopt temperature scaling, a widely-used method for uncertainty calibration, which is also helpful in our method.

Compared with Sigmoid loss, Logistic loss is much smoother to smoothly add the unlabeled data into the training process of PU learning. On the contrary, sigmoid produces a ``step-wise'' learning type to add the samples. Moreover, the sigmoid function is bounded from above, while the logistic function is not. In the experiment section, we conduct extensive experiments on both functions and report the results.

\subsection{Sample Weighting}

In the second stage, we update the weight of each sample based on its hardness. To map the hardness of each sample to its weight dynamically, we propose two functions, i.e. the hardness-weight mapping function and the training scheduler.

\textbf{Hardness-weight mapping function.}
We adopt the minimizer function of SPL-IR-welsch~\cite{fan2017self} to map the hardness of each sample to its weight:
\begin{align}
    v_i^*=\exp(-\frac{d_i}{\lambda^2})\label{eq:wei}
\end{align}
where $v_i^*$ denotes the weight of the $i$-th sample, $d_i$ denotes the hardness of the $i$-th sample, and $\lambda$ denotes the threshold.

\textbf{Training Scheduler.}
In order to make the weights of the samples change as the training process proceeds, we introduce several training schedulers to control the threshold. These schedulers are defined by their corresponding pacing function.

Inspired by previous studies in curriculum learning~\cite{wang2019dynamic,wang2021survey}, we propose a set of pacing functions to deal with different scenarios. The detained formulation of these functions as well as the semantic explanation of these functions are introduced in the following.

The self-paced pacing function $F(t)$ returns a value monotonically increasing from $\lambda_0$ to $\beta$ with the input variable $t$, which represents the current training iteration. It monitors the model learning status and controls the curriculum learning speed. We explore several function classes as follows:

\begin{itemize}

    \item \texttt{Linear function}: indicating the constant learning speed:
    \begin{align}
        F_{linear} (t) = \min(\beta, \lambda_0+\frac{\beta-\lambda_0}{T_{grow}}\cdot t)\label{sf:linear}
    \end{align}
    where $\beta$ denotes the final value of the threshold, $\lambda_0$ denotes the initial value of the threshold, $T_{grow}$ denotes the iteration when the threshold reaches $\beta$ for the first time, $t$ denotes the current iteration.
    
    \item \texttt{Convex function}: indicating the learning speed from fast to slow:
    \begin{align}
        F_{convex}(t) = 
            \begin{cases}
                \lambda_0+(\beta-\lambda_0)*\sin(\frac{t}{T_{grow}}*\frac{\pi}{2}), &\text{if $t\le T_{grow}$}\\
                \beta, &\text{otherwise}\label{sf:convex}
            \end{cases}
    \end{align}
    
    
    \item \texttt{Concave function}: indicating the learning speed from slow to fast:
    \begin{align}
        F_{concave}(t) =  
            \begin{cases}
                \lambda_0+(\beta - \lambda_0) * (1 - \cos(\frac{t}{T_{grow}} * \frac{\pi}{2})), &\text{if $t\le T_{grow}$}\\
                \beta, &\text{otherwise}\label{sf:concave}
            \end{cases}
    \end{align}

    \item \texttt{Exponential function}: indicating an exponential change of the learning speed from fast to slow:
      \begin{align}
            F_{exp}(t) = \lambda_0 + (\beta - \lambda_0) * (1 - \gamma^t)\label{sf:exp}
        \end{align}
        where $\gamma$ is the hyper-parameter that controls the speed of change.
    
\end{itemize}
Different classes of pacing functions indicate different learning styles and can be used in different scenarios. In the experiment section, we apply these pacing functions on multiple datasets and report the results of \ourmethod equipped with different pacing functions.

\subsection{Weighted Supervised Training}

In the final stage, we treat the unlabeled data as negative and convert the original semi-supervised training into supervised training with weighted samples. In this stage, various supervised training methods and losses can be used. In our work, we adopt the binary cross-entropy loss for simplicity. Thus the supervised training loss for the $i$-th sample, i.e. $L_i$ in Eq. (\ref{equ:hardness}) is 
\begin{align}
    L(q_i,y_i)=-y_i\cdot\log q_i-(1-y_i)\cdot\log (1-q_i)\label{eq:bce}
\end{align}
where $q_i$ is the probability of the $i$-th sample being positive, given by the current model. $y_i$ is the ground truth label, note that $y_i=0$ for unlabeled samples.

Finally, in this stage, we update the model parameter by minimizing the loss of all weighted samples:
\begin{align}
    \min_{\mathbf{\theta}}\sum_{i=1}^{n_p}v_iL(q_i,1)+\sum_{i=1}^{n_u}v_iL(q_i,0)
\end{align}
where $v_i$ denotes the current weight of the $i$-th sample, $n_p$ is the number of positive samples, and $n_u$ is the number of unlabeled samples.

In the experiments, we find that performing multiple epochs in this stage is helpful in improving the performance of the final model. To this end, we propose to perform the weighted supervised training for $E$ epochs.

\subsection{Overall Objective}\label{sec:overall}

While one can implement our method by following our three-stage strategy, we also provide an overall objective function in Eq. (\ref{equ:hardness}) to help readers understand our method from a more holistic perspective. Note that the objective function we proposed in Eq. (\ref{equ:hardness}) contains two kinds of parameters, namely the classifier's parameters $\theta$ and the samples' weights $\mathbf{v}^p, \mathbf{v}^u$, and we alternately optimizes these two types of parameters in our strategy. Thus the whole framework of our method is equivalent to minimizing Eq. (\ref{equ:hardness}) via the alternative search strategy (ASS):
\begin{align}
    \min_{\mathbf{\theta},\mathbf{v}^p, \mathbf{v}^u}\mathbb{E}(\mathbf{\theta},\mathbf{v}^p, \mathbf{v}^u;\lambda^p, \lambda^u)&=\sum_{i=1}^{n_p} v^p_i L^+_i+g(v^p_i, d^p_i;\lambda^p)\label{equ:hardness}\\
    &+\sum_{i=1}^{n_u} v^u_i L^-_i+g(v^u_i,d^u_i;\lambda^u)\nonumber
\end{align}
where $d^p, d^u$ denote the hardness for distinguishing the easy and clean samples from the positive and unlabeled data, respectively. $v_i$ is the weight of the $i$-th sample. $L_i$ is the loss of the $i$-th sample in the ``Weighted Supervised Training'' stage in the previous iteration. $g(v,d;\lambda)$ is the implicit regularizer SPL-IR-welsch~\cite{fan2017self}. It is the dual potential function of the Welsch function $\lambda^2\left(1-\exp\left(-\frac{L^2}{\lambda^2}\right)\right)$, which is a well-studied robust function, and the minimizer function of $g(v,d;\lambda)$ is our hardness-weight mapping function in Eq. (\ref{eq:wei}).  It is worth noting that there is no analytic form of the regularizer $g$. However, we maintain the form in Eq. (\ref{equ:hardness}) as per the custom in self-paced learning. $\lambda^p$ and $\lambda^u$ are the thresholds for positive samples and unlabeled samples, respectively. We change these thresholds throughout the training process and derive them from the training scheduler.



\section{Experiments}
In this section, we empirically evaluate the performance of our method compared with the state-of-art approaches and justify the benefit of our method on commonly used benchmarks. We also investigate the impact of hyper-parameters and conduct ablation studies to analyze the effect of our method.
Due to space limitations, some experimental results are reported in Appendix~\ref{sec:appendix}.

\subsection{Experimental Setup}

\textbf{Datasets}

We perform experiments on eight commonly used public datasets: (1) \texttt{CIFAR-10}, (2) \texttt{STL-10}, (3) \texttt{MNIST}, (4) \texttt{F-MNIST}, (5) \texttt{Alzheimer} and (6) three datasets from UCI Machine Learning Repository, including \texttt{mushrooms}, \texttt{shuttle} and \texttt{spambase}. Since some datasets have multiple classes, we preprocess it as a binary classification problem following the processing of conventions~\cite{kato2018learning}. Specifically, we choose some of the classes as positive classes, and the others as negative classes, as shown in Table~\ref{tab:class_setting}.
The summary of datasets is listed in Table~\ref{tab:dataset}, including the number of instances, the feature dimensions, and the number of positive and negative instances.
\begin{table}[htbp]
    \caption{Preprocessing of the datasets.}
    \centering
\resizebox{1.\linewidth}{!}{
    \begin{tabular}{ccc}\toprule
         Dataset & Positive class & Negative class\\
         \midrule
         \texttt{CIFAR-10} & airplane, truck, automobile, ship & bird, cat, deer, dog, frog, horse\\
         \texttt{STL-10} & 0, 2, 3, 8, 9 & 1, 4, 5, 6, 7\\
         \texttt{MNIST} & 0, 2, 4, 6, 8 & 1, 3, 5, 7, 9\\
         \texttt{F-MNIST} & 0, 2, 4, 6 & 1, 3, 5, 7, 8, 9\\
         \texttt{shuttle} & 1 & 2, 3, 4, 5, 6, 7\\
         \bottomrule
    \end{tabular}
    }
    \label{tab:class_setting}
\end{table}

\begin{table}[htbp]
    \caption{Details of the datasets in our experiments.}
    \centering
    \begin{tabular}{ccccc}\toprule
         Dataset & \#Ins & \#Fea & \#Pos & \#Neg\\
         \midrule
         \texttt{CIFAR-10} & 60,000 & 3 × 32 × 32 & 24,000 & 36,000\\
         \texttt{STL-10} & 13,000 & 3 × 96 × 96 & 6,500 & 6,500\\
         \texttt{MNIST} & 70,000 & 28 × 28 & 34,418 & 35,582\\
         \texttt{F-MNIST} & 70,000 & 28 × 28 & 28,000 & 42,000\\
         \texttt{Alzheimer} & 6400 & 3 × 224 × 224 & 960 & 5,440\\
         \texttt{mushroom} & 8,124 & 22 & 3,912 & 4,208\\
         \texttt{shuttle} & 58,000 & 9 & 12,414 & 45,586\\
         \texttt{spambase} & 4,601 & 57 & 1,813 & 2,788\\
         \bottomrule
    \end{tabular}
    \label{tab:dataset}
\end{table}

\noindent\textbf{Compared Methods}

To demonstrate the superiority of \ourmethod, we compare it with \textbf{PN} baseline which simply regards unlabeled data as negative, together with seven best-perform PU learning algorithms to date: 1) \textbf{uPU}~\cite{du2014analysis} introduces a general unbiased risk estimator. 2) \textbf{nnPU}~\cite{kiryo2017positive} proposes a non-negative risk estimator for PU learning and fixes the overfitting problem in PU learning. 3) \textbf{PUSB}~\cite{kato2018learning} captures the existence of a selection bias in the labeling process and achieves compatible performance on some benchmarks. 4) \textbf{PUbN}~\cite{hsieh2019classification} proposes to incorporate biased negative data based on empirical risk minimization and shows high performance on \texttt{CIFAR-10} and \texttt{MNIST}. 5) \textbf{PULNS}~\cite{luo2021pulns} is a powerful PU learning approach that selects effective negative samples from unlabeled data based on reinforcement learning. 6) \textbf{P$^3$Mix}~\cite{li2022your} adopt the mixup technique and select the mixup partners for marginal pseudo-negative instances from the positive instances that are around the learned boundary. For each setting, we show the best result among P$^3$Mix-E and P$^3$Mix-C. 7) \textbf{Dist-PU}~\cite{zhao2022dist} proposes to pursue the label distribution consistency between predicted and ground-truth label distribution and achieves state-of-the-art performance on several benchmarks. All hyperparameters of the compared methods are tuned in the same way as reported in the references.

\noindent\textbf{Evaluation Metric}

We select the error rate of classification (i.e., 1 - accuracy) on the testing dataset as our evaluation metric, which is consistent with the common practice in PU learning. For each configuration of our method and all compared methods, we conduct 10 independent runs and report the average error rate and standard deviation.

\begin{table*}[htbp]
    \caption{Classification errors of nine methods on seven datasets with $\pi\in\{0.2, 0.4, 0.6\}$. The best values are \textbf{bold}. The average error rate of classification in test data (\%) and standard deviation over 10 trials are reported.}
    \label{tab:all_res}
    \centering
    \begin{tabular}{ccccccccc} \toprule
        $\pi$ & Method & CIFAR-10 & STL-10 & MNIST & F-MNIST & mushrooms & shuttle & spambase\\
        \midrule
        \multirow{9}{*}{0.2} & \texttt{PN} & 17.98 (0.016) & 16.02 (0.012) & 14.02 (0.004) & 5.20 (0.004) & 1.70 (0.004) & 7.17 (0.007) & 17.37 (0.010)\\
        & \texttt{uPU} & 11.60 (0.028) & 13.21 (0.023) & 8.87 (0.006) & 4.12 (0.003) & 1.59 (0.007) & 4.03 (0.011) & 10.80 (0.019)\\
        & \texttt{nnPU} & 12.63 (0.008) & 14.63 (0.032) & 7.38 (0.009) & 4.21 (0.006) & 1.42 (0.008) & 3.06 (0.009) & 10.23 (0.004)\\
        & \texttt{PUSB} & 11.60 (0.010) & 14.29 (0.025) & 6.24 (0.005) & 4.10 (0.005) & 1.58 (0.006) & 3.08 (0.009) & 10.08 (0.005)\\
        & \texttt{PUbN} & 9.24 (0.009) & 13.21 (0.026) & 5.87 (0.006) & 4.22 (0.005) & 1.78 (0.005) & 2.36 (0.005) & 9.42 (0.020)\\
        & \texttt{PULNS} & 8.47 (0.009) & 12.16 (0.015) & 3.64 (0.009) & 4.00 (0.004) & 0.40 (0.004) & 1.88 (0.008) & 8.71 (0.020)\\
        & \texttt{P$^3$Mix} & 7.99 (0.009) & 10.92 (0.015) & 3.61 (0.007) & 3.82 (0.003) & 0.37 (0.003) & 1.76 (0.005) & 8.03 (0.007)\\
        & \texttt{Dist-PU} & 8.68 (0.008) & 9.99 (0.008) & 3.54 (0.003) & 3.90 (0.004) & 0.34 (0.003) & 1.79 (0.006) & 8.01 (0.011)\\
        & \texttt{ours} & \textbf{7.58} (0.004) & \textbf{9.95} (0.012) & \textbf{3.28} (0.003) & \textbf{3.59} (0.003) & \textbf{0.24} (0.002) & \textbf{1.23} (0.004) & \textbf{6.40} (0.006)\\
        \hline
        \multirow{9}{*}{0.4} & \texttt{PN} & 21.71 (0.016) & 26.48 (0.017) & 17.27 (0.011) & 8.08 (0.017) & 4.78 (0.017) & 8.83 (0.014) & 22.09 (0.032)\\
        & \texttt{uPU} & 13.61 (0.012) & 20.73 (0.016) & 12.90 (0.010) & 5.44 (0.007) & 1.95 (0.012) & 5.31 (0.007) & 15.96 (0.039)\\
        & \texttt{nnPU} & 12.60 (0.013) & 16.15 (0.027) & 7.74 (0.008) & 5.03 (0.002) & 1.77 (0.009) & 5.14 (0.010) & 15.77 (0.031)\\
        & \texttt{PUSB} & 12.62 (0.011) & 17.47 (0.031) & 8.09 (0.014) & 4.97 (0.005) & 1.99 (0.011) & 3.69 (0.012) & 13.16 (0.021)\\
        & \texttt{PUbN} & 11.13 (0.018) & 16.89 (0.018) & 7.00 (0.020) & 4.98 (0.003) & 1.82 (0.011) & 4.44 (0.010) & 12.86 (0.032)\\
        & \texttt{PULNS} & 10.76 (0.010) & 16.98 (0.030) & 5.62 (0.006) & 4.76 (0.004) & 1.32 (0.009) & 2.79 (0.013) & 12.03 (0.021)\\
        & \texttt{P$^3$Mix} & 10.52 (0.007) & 16.28 (0.020) & 4.34 (0.004) & 4.69 (0.005) & 1.29 (0.011) & 2.90 (0.010) & 11.89 (0.024)\\
        & \texttt{Dist-PU} & 10.35 (0.006) & 16.91 (0.022) & 4.36 (0.004) & 4.68 (0.004) & 1.29 (0.006) & 3.08 (0.012) & 11.85 (0.024)\\
        & \texttt{ours} & \textbf{10.26} (0.006) & \textbf{15.62} (0.017) & \textbf{3.80} (0.003) & \textbf{4.56} (0.003) & \textbf{0.61} (0.004) & \textbf{1.36} (0.003) & \textbf{9.79} (0.014)\\
        \hline
        \multirow{9}{*}{0.6} & \texttt{PN} & 31.39 (0.044) & 37.15 (0.055) & 21.48 (0.021) & 17.81 (0.036) & 19.61 (0.042) & 24.42 (0.052)& 34.33 (0.026)\\
        & \texttt{uPU} & 14.58 (0.012) & 23.38 (0.032) & 16.17 (0.010) & 8.50 (0.009) & 2.02 (0.009) & 5.32 (0.009) & 25.31 (0.029)\\
        & \texttt{nnPU} & 13.85 (0.014) & 19.06 (0.012) & 9.85 (0.008) & 8.03 (0.002) & 2.06 (0.015) & 5.27 (0.009) & 22.43 (0.032)\\
        & \texttt{PUSB} & 12.23 (0.015) & 20.60 (0.019) & 8.22 (0.018) & 7.26 (0.007) & 3.09 (0.013) & 3.91 (0.011) & 15.39 (0.031)\\
        & \texttt{PUbN} & 11.62 (0.010) & 19.01 (0.023) & 7.62 (0.014) & 7.63 (0.003) & 2.36 (0.011) & 3.72 (0.020) & 16.18 (0.030)\\
        & \texttt{PULNS} & 11.39 (0.014) & 20.03 (0.022) & 5.81 (0.004) & 5.28 (0.004) & 2.01 (0.004) & 3.64 (0.014) & 13.54 (0.025)\\
        & \texttt{P$^3$Mix} & 10.88 (0.010) & 19.32 (0.015) & 5.19 (0.006) & 4.85 (0.005) & 1.72 (0.005) & 3.57 (0.012) & 13.54 (0.025)\\
        & \texttt{Dist-PU} & 10.77 (0.007) & 19.30 (0.027) & 5.22 (0.005) & 4.62 (0.005) & 1.31 (0.005) & 3.52 (0.010) & 13.55 (0.023)\\
        & \texttt{ours} & \textbf{10.73} (0.005) & \textbf{18.15} (0.011) & \textbf{4.48} (0.003) & \textbf{4.39} (0.002) & \textbf{0.84} (0.003) & \textbf{1.50} (0.006) & \textbf{11.25} (0.009)\\
        \bottomrule
    \end{tabular}

\end{table*}



\noindent\textbf{Dataset construction}

We construct positive-unlabeled datasets for each binary-labeled dataset. 

As for the training dataset, we divide it into $\mathcal{X}_p$ consisting of $n_p$ positive data which is sampled randomly from all positive samples in the original dataset, and $\mathcal{X}_u$ consisting of $n_u$ unlabeled data. To construct $\mathcal{X}_u$, we introduce $\pi$ as the proportion of positive samples in $\mathcal{X}_u$, and randomly select $n_u\cdot \pi$ positive samples and $n_u\cdot (1-\pi)$ negative samples from the corresponding distribution in the original dataset. We set the value of $n_p$ and $n_u$ following the setup in PUSB~\cite{kato2018learning}. Specifically, for \texttt{CIFAR-10}, \texttt{F-MNIST} and \texttt{MNIST}, we set $n_p=2000$ and $n_u=4000$. For \texttt{STL-10}, we set $n_p=1000$ and $n_u=2000$. For \texttt{Alzheimer}, \texttt{mushroom}, \texttt{shuttle} and \texttt{spambase}, we set $n_p=400$ and $n_u=800$. For all datasets, in order to demonstrate different scenarios, we select the class prior (i.e. $\pi$) from $\{0.2, 0.4, 0.6\}$, following the setup in PULNS~\cite{luo2021pulns}.

As for the validation dataset and the testing dataset, we generate them in the same way as $\mathcal{X}_u$ in the training dataset along with the same setting of values for $\pi$. We use $|Val|$ and $|Test|$ to denote the number of samples in the validation dataset and testing dataset. For \texttt{CIFAR-10}, \texttt{STL-10}, \texttt{F-MNIST} and \texttt{MNIST}, we set $|Val|=500$ and $|Test|=5000$. For \texttt{Alzheimer}, \texttt{mushroom}, \texttt{shuttle} and \texttt{spambase}, we set $|Val|=100$ and $|Test|=1000$.

\noindent\textbf{Classifier Specification}

We use different classifiers for different datasets, following the setting in previous work~\cite{kato2018learning,luo2021pulns}. For \texttt{CIFAR-10} and \texttt{STL-10}, we adopt an all-convolutional neural network. For \texttt{Alzheimer}, we use ResNet-50. For the remaining datasets, we adopt a multilayer perceptron (MLP) with ReLU activation and a single-hidden-layer of 100 neurons.

\noindent\textbf{Hyperparameter Setting}

We pre-train our classifiers with nnPU for 100 epochs and train for 20 epochs in each iteration. We apply the Adam optimizer and set the batch size to 64. The learning rate is tuned from $\{10^{-2}, 10^{-3}, 10^{-4}\}$. The weight decay is tuned from $\{0, 10^{-4}, 10^{-2}\}$, and the number of warm-up epochs in each iteration is tuned from \{0, 5, 10\}. We tune the temperature $\tau$ from $[0.2, 3.0]$, initial threshold $\lambda_0$ from $[0.1, 0.9]$, maximum threshold $\beta$ from $\{1, 2, 5\}$, epoch number $E$ from $\{10, 20, 50\}$, and number of growing steps $T_{grow}$ from $\{10, 15, 20\}$. We conduct early stopping based on the accuracy on the validation set and stop iterating when the accuracy on the validation set doesn't increase for 5 consecutive iterations.

\subsection{Experimental Results}

\textbf{Comparison with state-of-the-art methods.}
We conduct extensive experiments on eight datasets to demonstrate the superiority of our method against compared methods. The experimental results on seven datasets are listed in Table~\ref{tab:all_res}. And we show the experimental results on a more challenging dataset, i.e., the Alzheimer’s Dataset in Table~\ref{tab:alzheimer}. Note that in all compared methods, class-prior i.e. $\pi$ is assumed to be known, while it is not required in our method.

Table~\ref{tab:all_res} summarizes the performance of all methods on seven datasets under three values of $\pi$ in terms of average error rate and standard deviation over ten trials. From the results, we see that Dist-PU and P$^3$Mix are the best baselines on all datasets. However, neither of them is consistently better than the other. In contrast, our proposed \ourmethod outperforms all the compared methods by a significant margin on all datasets and achieves the best performance in all cases. This validates the effectiveness of our proposed method. Specifically, on dataset \texttt{spambase} with $\pi=$ 0.2, 0.4 and 0.6, \ourmethod achieves the average error rates of 6.40\%, 9.79\%, and 11.25\% respectively, while the best baseline gets 8.01\%, 11.85\%, and 13.54\% respectively. Also note that \ourmethod consistently achieves the lowest standard deviation in almost all cases, which verifies the stability of our proposed method. Moreover, recent works suffer from a significant performance drop when $\pi$ increases, as results on \texttt{shuttle} with different $\pi$ show. In contrast, the performance of \ourmethod only shows a slight drop and maintains state-of-the-art performance even when $\pi$ increases. Since we take the positive samples in the unlabeled data as the noise in negative data, the results validate the robustness of \ourmethod under large noise level.

\begin{figure*}[htbp]
\centering
  \subfigure[\ourmethod]{
    \includegraphics[width=0.72\linewidth]{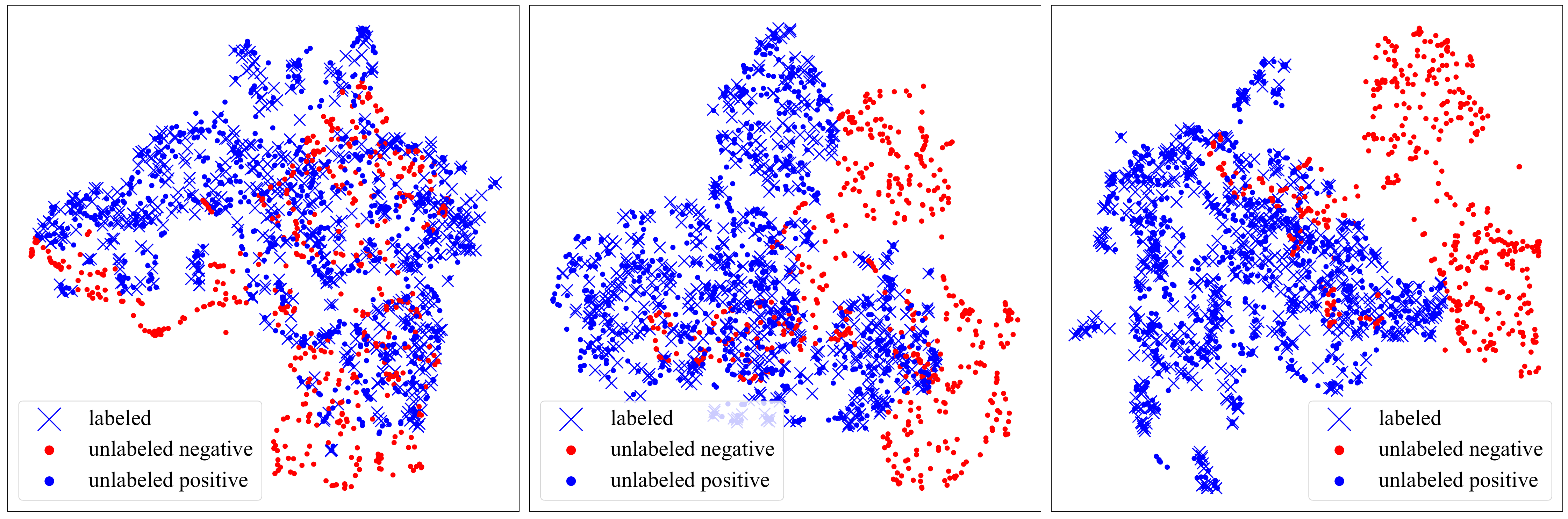}
  }
  \subfigure[Dist-PU]{
    \includegraphics[width=0.24\linewidth]{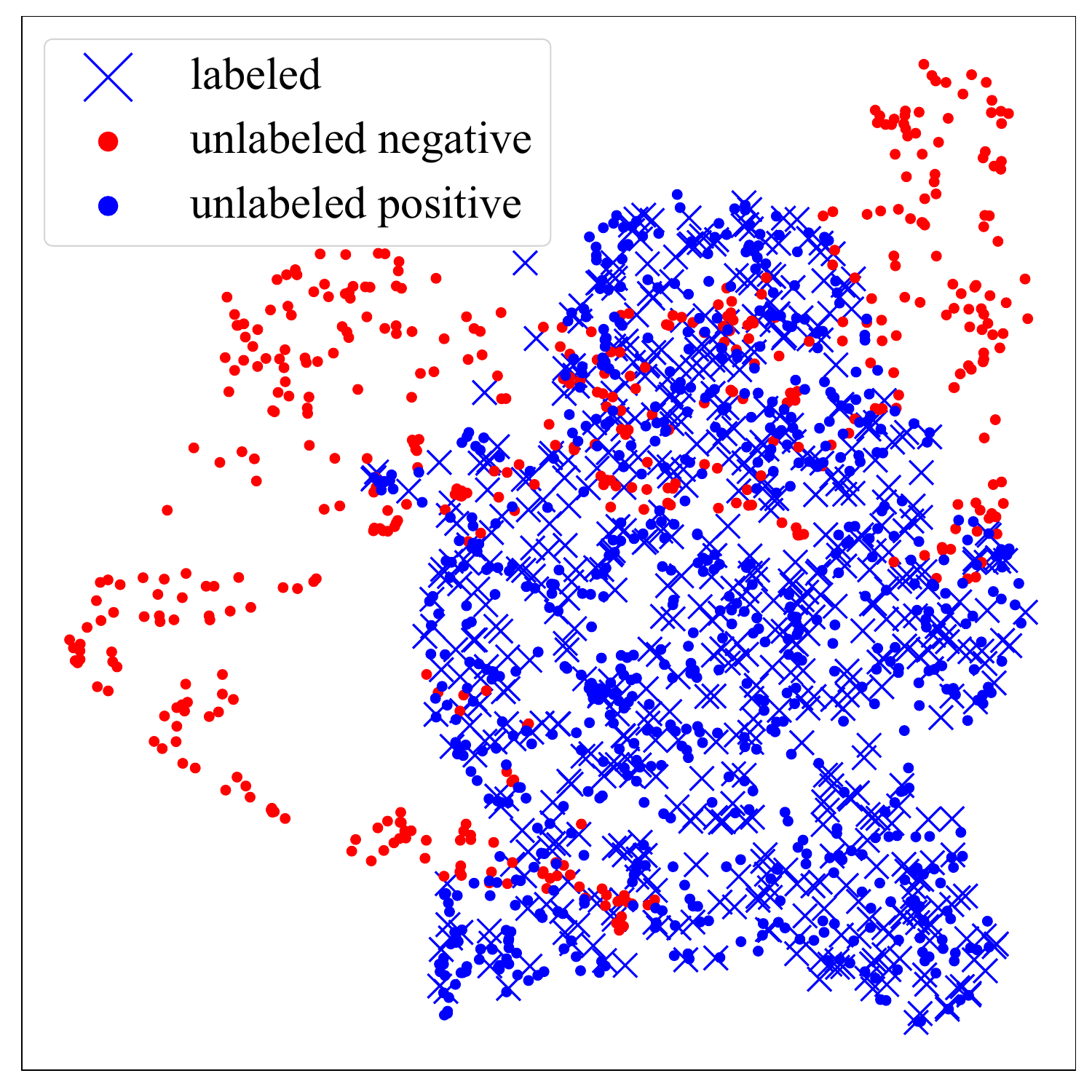}
  }
  \caption{UMAP visualization of the representations learned by (a) \ourmethod in the last three iterations and (b) Dist-PU on MNIST with $\pi=0.6$. Positive samples are blue and negative samples are red. 'x' represents labeled data and 'o' represents unlabeled data.}
  \label{fig:vis}
\end{figure*}

\begin{table}[htbp]
    \caption{Classification errors of \ourmethod on Alzheimer's Dataset.}
    \centering
  \resizebox{.85\linewidth}{!}{
    \begin{tabular}{cccc} \toprule
        Method & $\pi=0.2$ & $\pi=0.4$ & $\pi=0.6$ \\
        \midrule
        \texttt{uPU} & 31.22 (0.010) & 34.96 (0.016) & 35.88 (0.023) \\
        \texttt{nnPU} & 31.57 (0.012) & 34.46 (0.017) & 35.89 (0.027) \\
        \texttt{PUSB} & 30.97 (0.010) & 34.19 (0.018) & 34.96 (0.024) \\
        \texttt{PUbN} & 29.82 (0.008) & 32.99 (0.015) & 34.89 (0.022) \\
        \texttt{PULNS} & 29.47 (0.011) & 32.75 (0.014) & 33.90 (0.019) \\
        \texttt{Dist-PU} & 29.13 (0.008) & 31.97 (0.015) & 33.16 (0.021) \\
        \texttt{ours} & \textbf{28.49} (0.008) & \textbf{30.69} (0.014) & \textbf{32.71} (0.021) \\
        \bottomrule
    \end{tabular}
    }
    \label{tab:alzheimer}
\end{table}

\textbf{Results on Alzheimer's Dataset.}
To show the superiority of our method in more difficult real-world scenarios. We conduct experiments on the Alzheimer’s dataset~\cite{zhao2022dist,chen2020self}. We show the experimental results on Alzheimer's dataset over 5 random seeds in Table~\ref{tab:alzheimer}. On this dataset, we observe the same phenomena as in other datasets. That is, our method achieves a significantly lower classification error rate than the other methods. And we are able to have more minor performance degradation in scenarios where a larger $\pi$ is set, which verifies the robustness of our method.

\textbf{Visualization.}
In Figure~\ref{fig:vis}, we visualize the representations learned by  \ourmethod in the last three iterations and that learned by Dist-PU using UMAP~\cite{mcinnes2018umap}. Positive samples are blue and negative samples are red. 'x' represents labeled data and 'o' represents unlabeled data. We conduct the experiment on the \texttt{MNIST} dataset with $\pi=0.6$. We observe that 1) compared with the best baseline Dist-PU, \ourmethod learns much more distinguishable representations in the last iteration, which demonstrates the excellent classification ability of our proposed method. 2) As the training process proceeds, the boundary between positive and negative samples is becoming clearer, which verifies that our approach promotes the further learning of the classifier. 3) There is still some overlap between positive and negative samples, however, it may be beneficial to avoid overfitting.

\begin{figure*}[htbp]
\centering
  \subfigure[Dist-PU]{
    \includegraphics[width=0.48\linewidth]{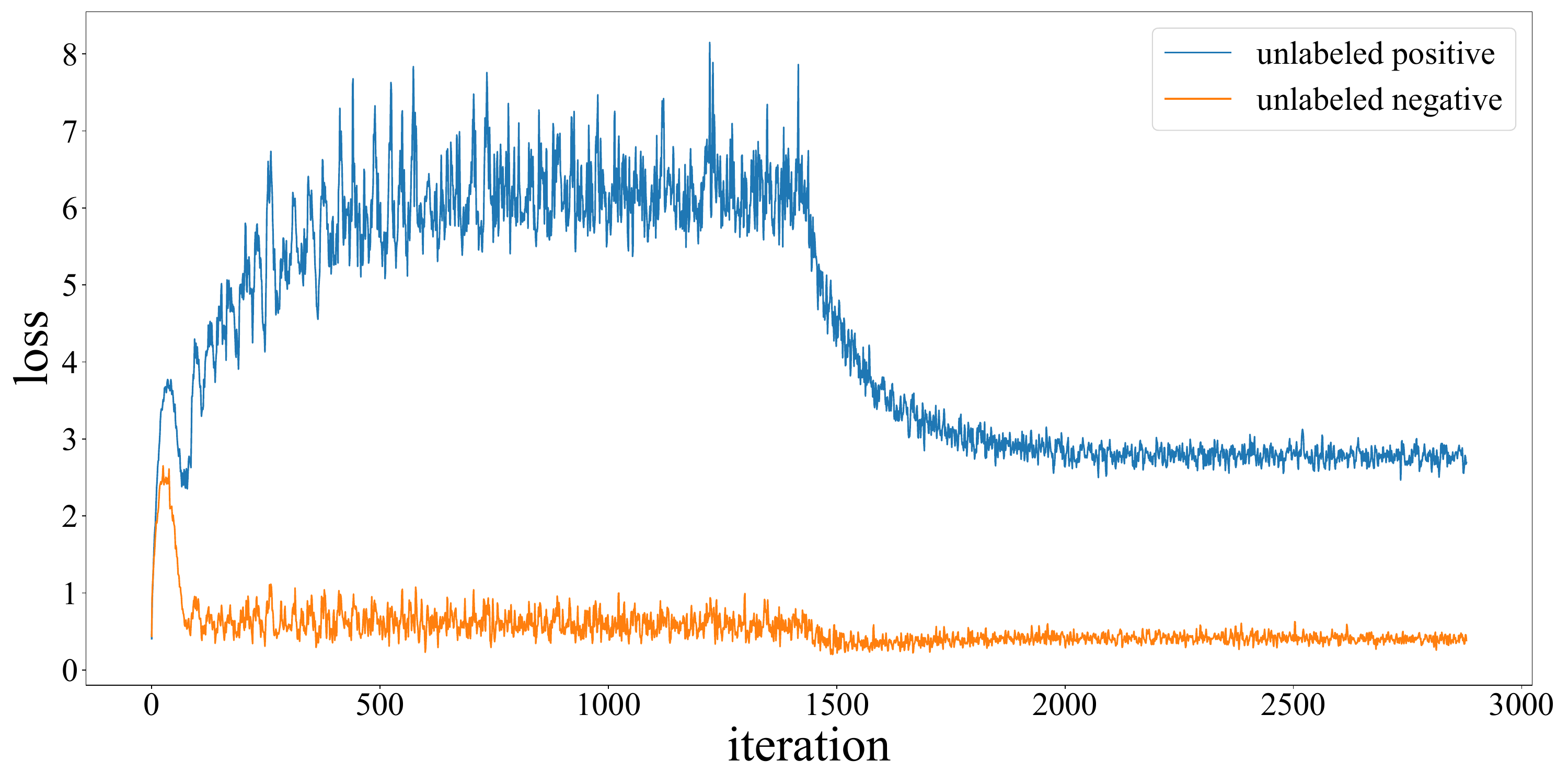}
  }
  \subfigure[\ourmethod]{
    \includegraphics[width=0.48\linewidth]{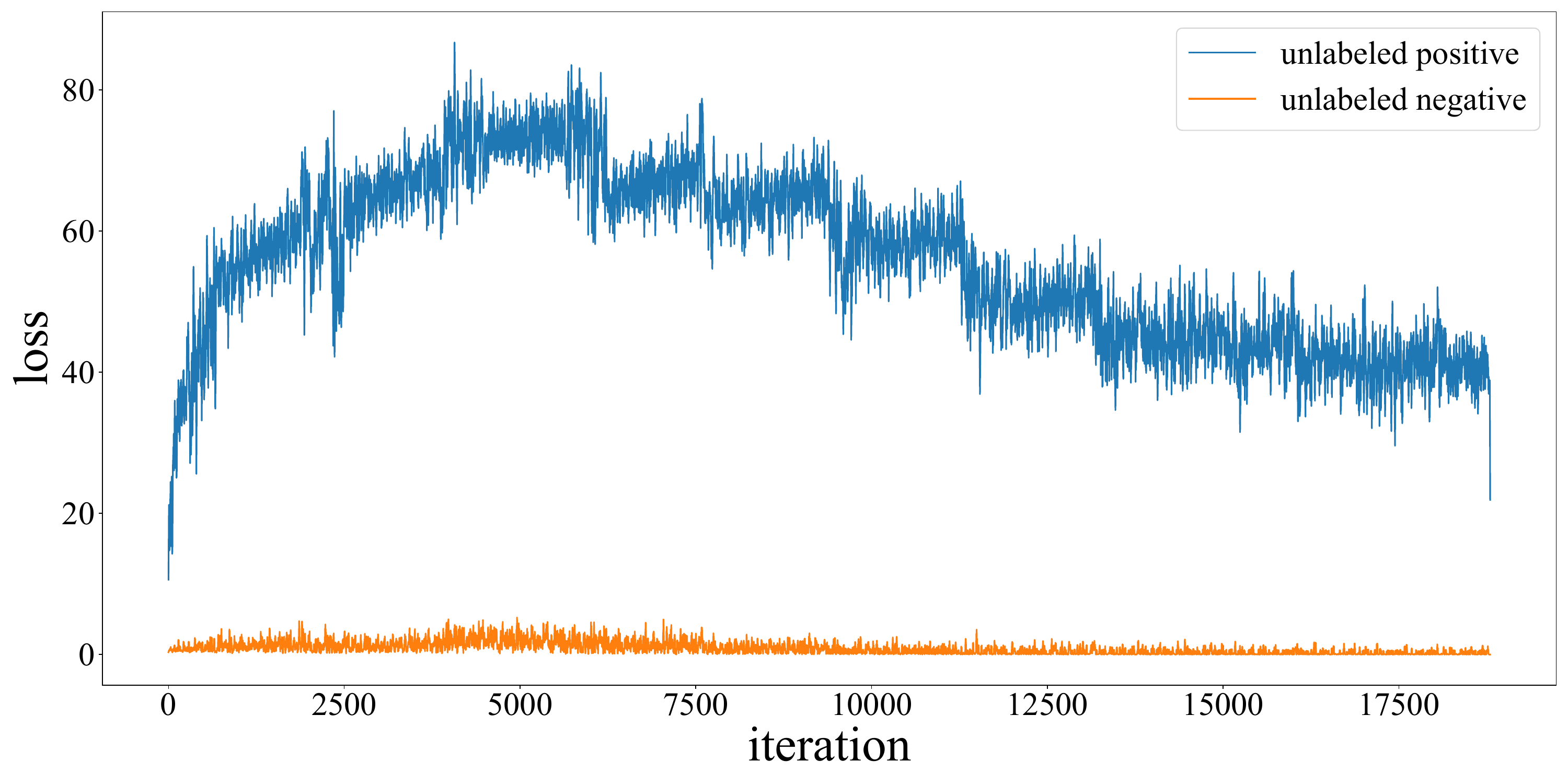}
  }
  \vspace{-.5em}
  \caption{Binary cross-entropy loss of (a) Dist-PU and (b) \ourmethod.}
  \label{fig:loss}
\end{figure*}

\textbf{Results on Unlabeled Samples.}
To validate the classification ability of our method on unlabeled data, we visualize the classification loss of unlabeled samples throughout the training process in Figure~\ref{fig:loss}. 1) We can observe that both Dist-PU and \ourmethod can fit the negative samples in the unlabeled samples quickly and achieve small losses. 2) However, there is a huge difference between the y-axis scales of the two graphs. For positive samples in the unlabeled samples (which are regarded as noises), our method always gives a much larger loss than that given by Dist-PU. This shows that our method can effectively distinguish noises in the unlabeled data and thus avoid over-fitting the noises. 3) Furthermore, we observe that after using entropy minimization and Mixup regularization, there appears an extremely rapid decrease in the loss on unlabeled positive samples. This may be due to the fact that regularization can polarize the predicted values, thus driving the model to misclassify the ``hard noises''. On the contrary, our method is able to dynamically change the weights of samples so that these noises are gradually removed and thus avoids persistent bias caused by incorrect pseudo labels.

\subsection{Ablation Studies}
In this section, we present the results of ablation studies to show the effect of each component in \ourmethod.

\begin{table}[htbp]
    \caption{Classification errors of \ourmethod with different hardness-weight mapping functions.}
    \centering
  \resizebox{\linewidth}{!}{
    \begin{tabular}{ccccc} \toprule
        & & $\pi=0.2$ & $\pi=0.4$ & $\pi=0.6$\\
        \midrule
        \multirow{3}{*}{CIFAR-10} & Hard & 8.656 (0.007) & 11.016 (0.008) & 11.436 (0.005)\\
        & Linear & 9.368 (0.010) & 11.176 (0.010) & 11.088 (0.007)\\
        & ours & \textbf{7.586} (0.004) & \textbf{10.260} (0.006) & \textbf{10.738} (0.005)\\
        \hline
        \multirow{3}{*}{MNIST} & Hard & 5.324 (0.007) & 7.562 (0.006) & 6.480 (0.006)\\
        & Linear & 4.608 (0.003) & 6.130 (0.005) & 8.972 (0.008)\\
        & ours & \textbf{3.282} (0.003) & \textbf{3.802} (0.003) & \textbf{4.482} (0.003)\\
        \hline
        \multirow{3}{*}{mushroom} & Hard & 0.610 (0.005) & 1.250 (0.008) & 2.280 (0.007)\\
        & Linear & 0.650 (0.003) & 1.610 (0.012) & 2.230 (0.006)\\
        & ours & \textbf{0.240} (0.002) & \textbf{0.610} (0.004) & \textbf{0.840} (0.003)\\
        \hline
        \multirow{3}{*}{shuttle} & Hard & 2.250 (0.014) & 3.180 (0.011) & 6.590 (0.005)\\
        & Linear & 1.940 (0.008) & 3.850 (0.008) & 2.840 (0.008)\\
        & ours & \textbf{1.230} (0.004) & \textbf{1.360} (0.003) & \textbf{1.500} (0.006)\\
        \hline
        \multirow{3}{*}{spambase} & Hard &8.850 (0.010) & 11.030 (0.016) & 12.490 (0.011)\\
        & Linear & 8.790 (0.012) & 12.080 (0.011) & 13.840 (0.015)\\
        & ours & \textbf{6.400} (0.006) & \textbf{9.790} (0.014) & \textbf{11.250} (0.009)\\
        \bottomrule
    \end{tabular}
    }
    \label{tab:abl_reg}
\end{table}

\textbf{Effect of hardness-weight mapping function.}
The hardness-weight mapping function learning plays an important role in our method. To validate its effectiveness, we compare the performance of \ourmethod equipped with different hardness-weight mapping functions. We use Hard~\cite{kumar2010self} and Linear~\cite{jiang2014easy} as the baseline and summarize the results of baselines and our method in Table~\ref{tab:abl_reg}. Hard simply set the sample weight as 1 if its hardness is smaller than the threshold, otherwise 0. Linear maps the hardness linearly to the weight.

From the results, We observe that our method, which uses the minimizer function of SPL-IR-Welsh as the hardness-weight mapping function, performs much better than the baselines. Note that different hardness-weight mapping functions correspond to different kinds of regularizers $g(v,d;\lambda)$ in Eq. (\ref{equ:hardness}). Both Hard and Linear correspond to the explicit form of $g(v,d;\lambda)$, while for SPL-IR-Welsh, $v\cdot L_+g(v,d;\lambda)$ corresponds to the Welsch function, which is a well-studied robust loss function, and the analytic form of $g(v,d;\lambda)$ is unknown. So the results are also consistent with the previous studies~\cite{wang2021survey} in curriculum learning that implicit SP-regularizers perform better than other types. Moreover, as the class prior increases, both Hard and Linear suffer from a significant performance drop, while the performance of \ourmethod our method shows only a slight drop. We attribute this advantage to the robust function corresponding to SPL-IR-Welsh, which reduces the impact of the loss of noisy samples through the meaningful transformation of the original loss $L$. Surprisingly, we find that \ourmethod with Linear is not always better than Hard, which indicates that if not carefully designed, soft hardness-weight mapping functions are not always superior to hard ones.

\begin{table}[htbp]
    \caption{Classification errors of \ourmethod with different training schedulers.}
    \centering
  \resizebox{\linewidth}{!}{
    \begin{tabular}{ccccc} \toprule
        & & $\pi=0.2$ & $\pi=0.4$ & $\pi=0.6$\\
        \midrule
        \multirow{5}{*}{CIFAR-10} & const & 9.468 (0.005) & 12.012 (0.004) & 11.492 (0.009)\\
        \cline{2-5}
        & convex & 8.358 (0.005) & 10.540 (0.004) & 10.852 (0.006) \\
        & concave & 7.682 (0.005) & 10.532 (0.004) & 11.023 (0.007) \\
        & exponential & 7.818 (0.004) & 10.512 (0.005) & 10.811 (0.005) \\
        & linear & \textbf{7.586} (0.004) & \textbf{10.260} (0.006) & \textbf{10.738} (0.005)\\
        \hline
        \multirow{5}{*}{MNIST} & const & 5.244 (0.006) & 7.492 (0.004) & 8.724 (0.006)\\
        \cline{2-5}
        & convex & 3.526 (0.005) & 3.804 (0.003) & 5.292 (0.004) \\
        & concave & 3.681 (0.004) & 3.906 (0.004) & 5.319 (0.003) \\
        & exponential & 3.422 (0.005) & \textbf{3.793} (0.005) & 4.512 (0.003) \\
        & linear & \textbf{3.282} (0.003) & 3.802 (0.003) & \textbf{4.482} (0.003)\\
        \hline
        \multirow{5}{*}{mushroom} & const & 1.630 (0.005) & 1.650 (0.006) & 2.090 (0.010)\\
        \cline{2-5}
        & convex & 0.510 (0.004) & 0.930 (0.006) & 1.030 (0.003) \\
        & concave & 0.520 (0.006) & 0.850 (0.005) & 1.100 (0.004) \\
        & exponential & 0.480 (0.004) & 0.720 (0.005) & 0.970 (0.005) \\
        & linear & \textbf{0.240} (0.002) & \textbf{0.610} (0.004) & \textbf{0.840} (0.003)\\
        \hline
        \multirow{5}{*}{shuttle} & const & 3.850 (0.006) & 3.440 (0.009) & 3.660 (0.010)\\
        \cline{2-5}
        & convex & 2.470 (0.005) & 1.630 (0.004) & 1.810 (0.007) \\
        & concave & 2.590 (0.006) & 1.620 (0.006) & 1.800 (0.009) \\
        & exponential & 1.920 (0.006) & 1.580 (0.005) & 1.970 (0.008) \\
        & linear & \textbf{1.230} (0.004) & \textbf{1.360} (0.003) & \textbf{1.500} (0.006)\\
        \hline
        \multirow{5}{*}{spambase} & const & 9.660 (0.009) & 12.350 (0.012) & 15.200 (0.006)\\
        \cline{2-5}
        & convex & \textbf{6.380} (0.007) & 10.820 (0.010) & \textbf{10.800} (0.007) \\
        & concave & 8.640 (0.005) & 12.370 (0.012) & 14.490 (0.008) \\
        & exponential & 7.820 (0.006) & 9.920 (0.013) & 11.920 (0.008) \\
        & linear & 6.400 (0.006) & \textbf{9.790} (0.014) & 11.250 (0.009)\\
        \bottomrule
    \end{tabular}
    }
    \label{tab:abl_ts}
\end{table}

\textbf{Effect of training scheduler.}
Our main idea is to change the weights of samples gradually. To investigate the effectiveness of our method, we conduct experiments to show the results of \ourmethod with different training schedulers and report the results in Table~\ref{tab:abl_ts}. We use the constant scheduler as the baseline, which fixes the threshold throughout the training process.

From the results, we observe that all schedulers that return a changing threshold outperform the constant scheduler in all cases. Therefore, the results validate the effectiveness of our idea. As $\pi$ gets larger, the performance of the constant training scheduler suffers from a significant drop while others remain good performance. Note that since we regard the positive data in unlabeled samples as noise, the results show that \ourmethod is more robust than the baseline. Moreover, the results demonstrate that the dynamic threshold can better reflect the learning status of the current iteration, thus helping the hardness-weight mapping function to better distinguish between positive and negative samples in the unlabeled data. Besides, each of the four proposed schedulers has its own pros and cons. So it is better to choose the training scheduler according to the characteristic of the dataset.

\section{Conclusion}

In this work, drawing inspiration from human learning and the field of curriculum learning, we present a novel training strategy for positive-unlabeled (PU) learning, referred to as \ourmethod. \ourmethod converts the original semi-supervised learning into supervised learning with weighted samples and dynamically updates the sample weights based on the scheduler of the training process and the hardness of each sample. Specifically, \ourmethod employs an iterative training strategy and we divide each iteration into three stages. In the first stage, we calculate the hardness for both positive and unlabeled samples by treating all of the unlabeled samples as negative and taking the classification losses as the hardness. In the second stage, we use a hardness-weight mapping function to determine the sample weights based on their hardness and dynamic threshold that reflects the scheduler of the training process. In the final stage, we adopt conventional supervised training methods with weighted samples. By gradually updating the weights of unlabeled data, \ourmethod greatly avoids persistent bias caused by misclassification errors in the early stage of training. Extensive experiments validate that the proposed \ourmethod can not only avoid overfitting on noises in unlabeled samples but also achieve state-of-the-art performance.

In the future, we plan to explore more hardness metric functions and study the range of applications of these hardness metrics. Moreover, in our experiments, we observe that each type of training scheduler has its own pros and cons and get different performance on different datasets. Thus designing more training schedulers and combining the pacing function with the characteristic of the dataset is also an interesting direction for future work.

\begin{acks}
The work was supported in part by National Natural Science Foundation of China under Grant (No. 92270119 and 62072182).
\end{acks}

\bibliographystyle{ACM-Reference-Format}
\balance
\bibliography{sample-base}

\appendix

\section{Hyperparameter Analysis}\label{sec:appendix}

\subsection{Effect of hardness metric functions}

\begin{table}[htbp]
    \caption{Classification errors of \ourmethod with different hardness metric functions (i.e. sigmoid loss and logistic loss).}
    \centering
  \resizebox{\linewidth}{!}{
    \begin{tabular}{ccccc} \toprule
        & & $\pi=0.2$ & $\pi=0.4$ & $\pi=0.6$\\
        \midrule
        \multirow{2}{*}{CIFAR-10} & sigmoid & 8.170 (0.006) & 11.730 (0.006) & 12.356 (0.006)\\
        & logistic & \textbf{7.586} (0.004) & \textbf{10.260} (0.006) & \textbf{10.738} (0.005)\\
        \hline
        \multirow{2}{*}{MNIST} & sigmoid & 4.454 (0.003) & 5.260 (0.003) & 5.718 (0.003)\\
        & logistic & \textbf{3.282} (0.003) & \textbf{3.802} (0.003) & \textbf{4.482} (0.003)\\
        \hline
        \multirow{2}{*}{mushroom} & sigmoid & 0.300 (0.003) & 0.670 (0.004) & 1.040 (0.005)\\
        & logistic & \textbf{0.240} (0.002) & \textbf{0.610} (0.004) & \textbf{0.840} (0.003)\\
        \hline
        \multirow{2}{*}{shuttle} & sigmoid & \textbf{1.180} (0.004) & 2.140 (0.008) & 3.365 (0.009)\\
        & logistic & 1.230 (0.004) & \textbf{1.360} (0.003) & \textbf{1.500} (0.006)\\
        \hline
        \multirow{2}{*}{spambase} & sigmoid & 7.690 (0.013) & 9.980 (0.015) & 12.985 (0.018)\\
        & logistic & \textbf{6.400} (0.006) & \textbf{9.790} (0.014) & \textbf{11.250} (0.009)\\
        \bottomrule
    \end{tabular}
    }
    \label{tab:abl_loss}
\end{table}

To investigate the effect of the hardness metric functions, we compare the results of \ourmethod equipped with different hardness metric functions, including sigmoid loss and logistic loss. The results are reported in Table~\ref{tab:abl_loss}. According to the results, it is apparent that \ourmethod with logistic loss achieves a lower error rate in all cases, especially on large datasets i.e., \texttt{CIFAR-10} and \texttt{MNIST}. Note that we use the ``large-small loss'' trick to identify noises in unlabeled samples. However, the sigmoid loss is bounded from above, thus limiting the hardness of the samples which should have greater losses. Besides, the logistic loss is much smoother than the sigmoid function which helps us smoothly add the unlabeled data into the training process of PU learning.

\subsection{Effect of temperature $\tau$}

\begin{figure}[htbp]
    \centering
    \subfigure[mushrooms]{
        \includegraphics[width=\linewidth]{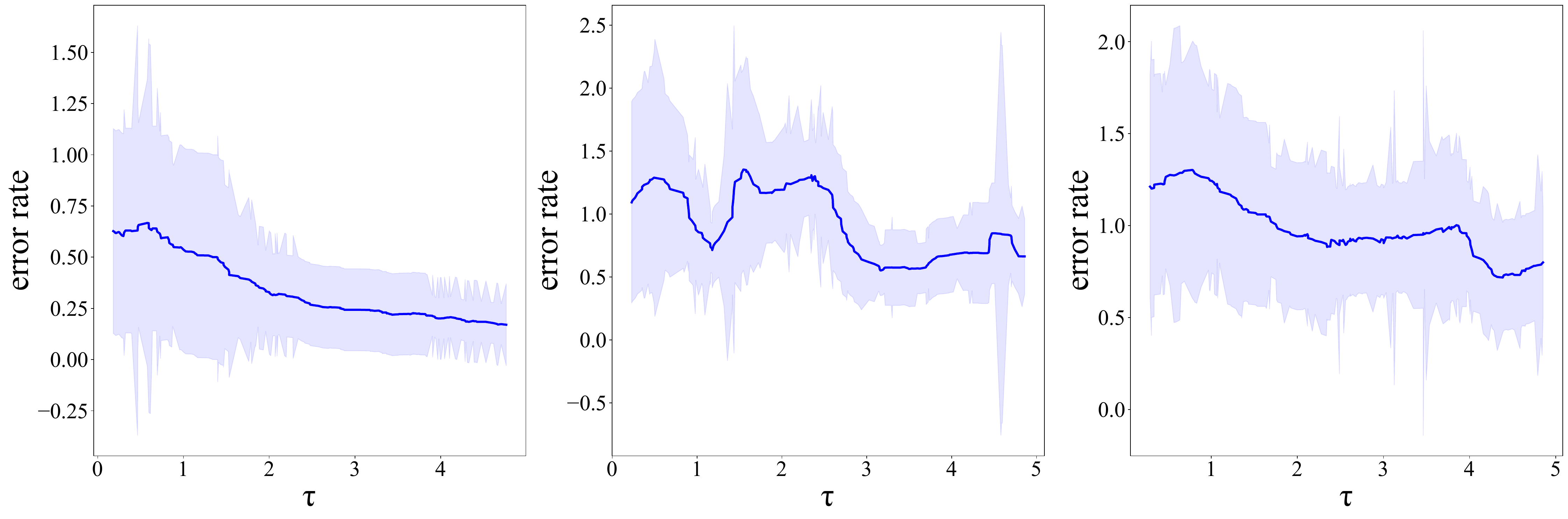}
        \label{fig:abl_tau_mush}
    }
    \quad
    \subfigure[CIFAR-10]{
    	\includegraphics[width=\linewidth]{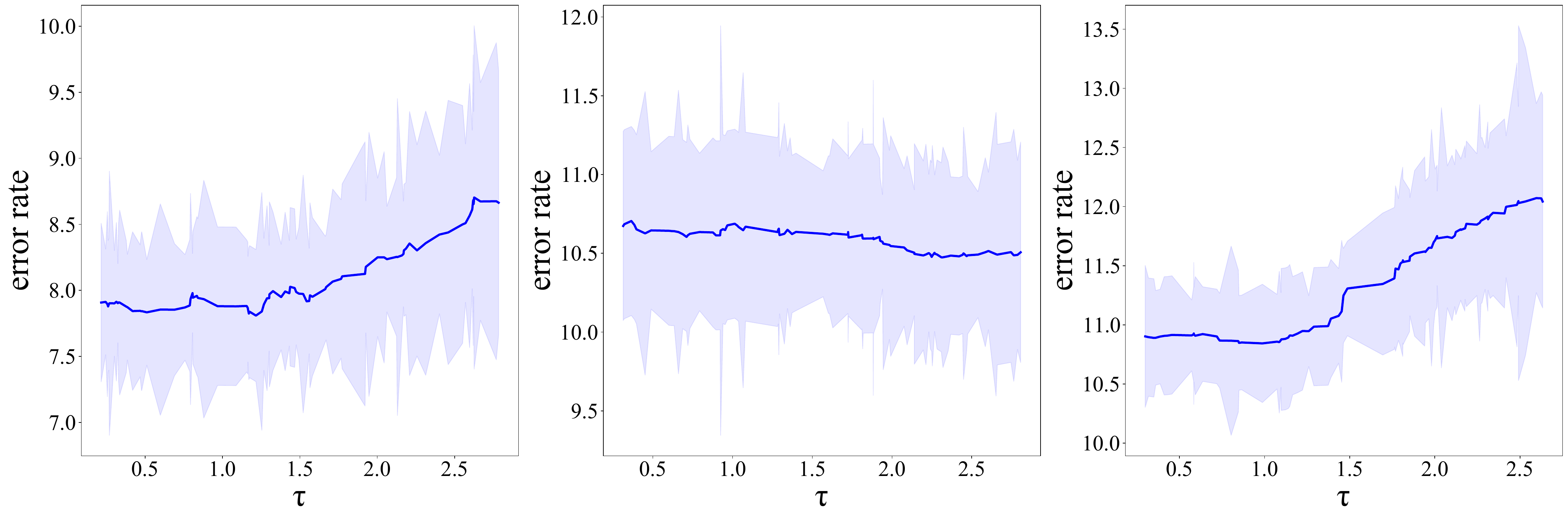}
    	\label{fig:abl_tau_cifar}
    }
    \caption{Curve of error rate (mean $\pm$ std) changing with $\tau$ on \texttt{mushroom} and \texttt{CIFAR-10} when $\pi=0.2$ (left), $0.4$ (middle) and $0.6$ (right).}
\end{figure}

Temperature scaling is a widely-used method for uncertainty calibration, and we find that it is also helpful in our method. We summarize the results of different settings of $\tau$ to quantify the effectiveness brought by $\tau$ in Eq. (\ref{eq:log}) and Eq. (\ref{eq:sig}). Specifically, in Figure~\ref{fig:abl_tau_mush} and Figure~\ref{fig:abl_tau_cifar}, we show the curve of error rate changing with $\tau$ on \texttt{mushroom} and \texttt{CIFAR-10}, respectively. From Figure~\ref{fig:abl_tau_mush}, we observe that the error rate decreases as $\tau$ increases. As a result, it is reasonable to select a large temperature (e.g. 4) on \texttt{mushroom}. From Figure~\ref{fig:abl_tau_cifar}, we observe that 1) the performance is stable when $\pi=0.4$ and 2) under the circumstance that  $\tau=0.2$ or $\tau=0.6$, the performance is stable as $\tau$ increases from $0.1$ to $1.0$, then starts dropping as $\tau$ continues to increase. As a result, it seems reasonable to set $\tau=1$ on \texttt{CIFAR-10}.

\subsection{Effect of epoch number $E$}

\begin{figure}[htbp]
    \centering
    \includegraphics[width=1.0\linewidth]{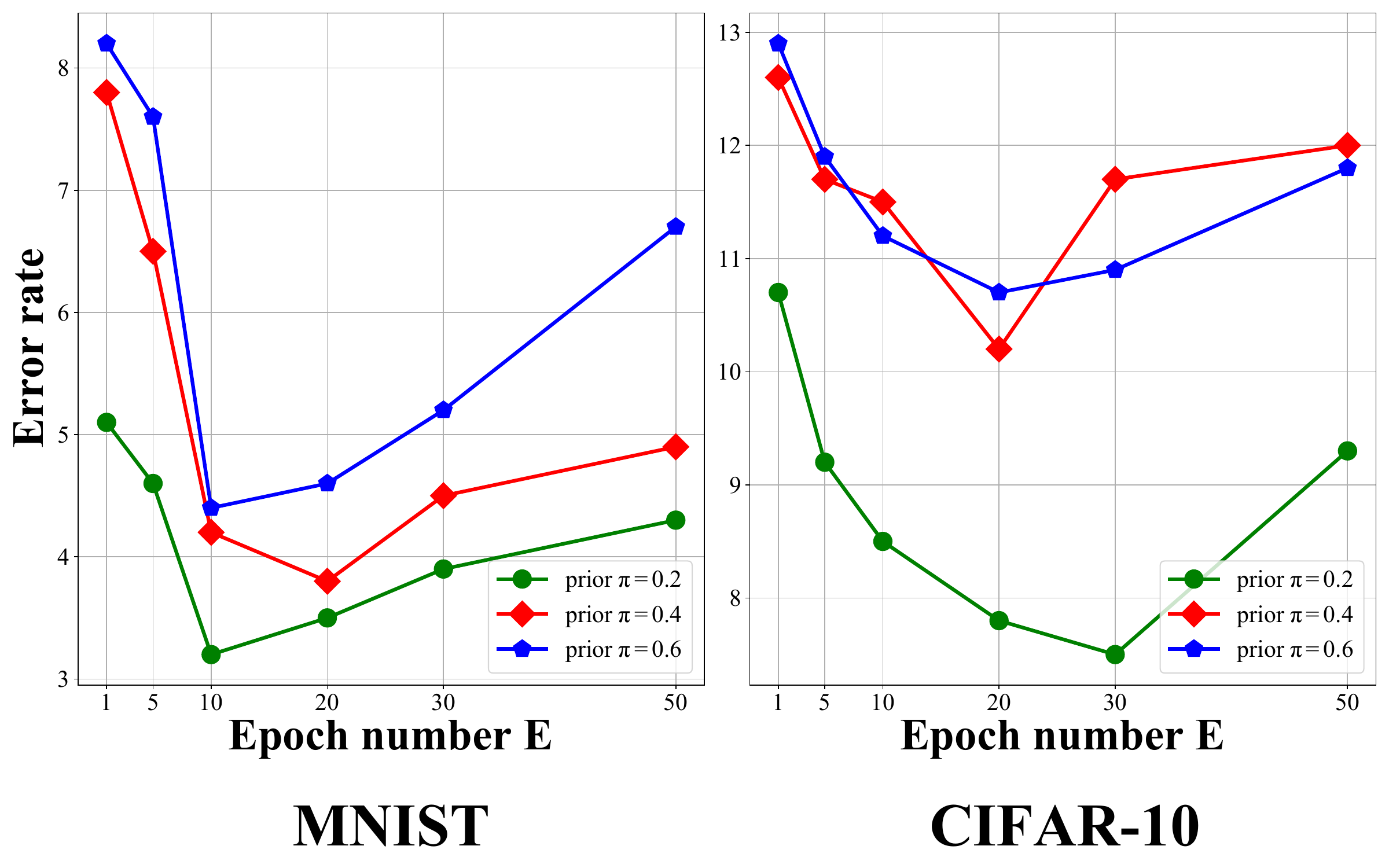}
    \caption{Classification error of \ourmethod with different epoch number. We conduct the experiments on MNIST (left) and CIFAR-10 (right) under different class priors.}
    \label{fig:epochs}
\end{figure}

In the final stage of each iteration, we convert the PU learning into supervised training and optimize the model for $E$ epochs. To study the role of the epoch number $E$, we conduct the experiments on MNIST and CIFAR-10 under different class priors and summarize the results of \ourmethod with different $E$ in Figure~\ref{fig:epochs}. From the results, we observe that: (1) The performance of \ourmethod becomes better as the number of epochs increases. It suggests that better model performance in the supervised training stage is needed to improve the final model. (2) However, it would lead to overfitting the misclassified samples and hurt the performance of the final model if the number of epochs is too large. (3) We find that setting the epoch number $E$ to 20 usually leads to good results on all datasets. Moreover, for larger datasets, we recommend using a relatively larger $E$, such as 30, if the training cost is acceptable because it is less likely to overfit on large datasets.

\end{document}